\documentclass{article}
\usepackage{acra}
\usepackage[T1]{fontenc}
\usepackage{mathptmx}            
\usepackage{graphicx}
\usepackage{booktabs}
\usepackage{placeins}
\usepackage{amsmath}
\usepackage{xcolor}
\usepackage{microtype}
\usepackage[font=small,labelfont=bf,labelsep=period,skip=4pt]{caption}
\usepackage[hidelinks]{hyperref}
\graphicspath{{figures/}}

\newcommand{\pck}[1]{PCK@#1}
\newcommand{\figref}[1]{Figure~\ref{#1}}
\newcommand{\tabref}[1]{Table~\ref{#1}}
\newcommand{\secref}[1]{Section~\ref{#1}}

\title{Real-Time Localisation of Dorsal Anatomical Regions on Dairy Cows\\in Commercial Rotary Parlours from RGB-D}
\author{Vineet Prasad, Shiron Melvin, Sheikh Azid and Hai Wang\thanks{Corresponding author: hai.wang@murdoch.edu.au}\\
School of Engineering and Energy, Murdoch University, Perth, Australia\\
\{vineet.vishaalprasad, shiron.antomelvin, sheikh.azid, hai.wang\}@murdoch.edu.au}

\begin{document}
\maketitle

\begin{abstract}
Automated monitoring of dairy cows increasingly measures one part of the body, such as the rump for temperature or the back for posture, yet the region itself is usually placed by hand. Existing systems either score the cow as a whole or detect sparse keypoints, often in controlled walkways, and neither yields the named region masks that such measurements need on a crowded commercial rotary platform. We present a perception front end that finds the target cow, ten bony landmarks and four named dorsal regions (withers, back, rump and tail-head) in one forward pass over a whole RGB-D frame. Colour and depth encoders are fused late and share one decoder with three heads. On an anatomical region dataset built on public rotary-parlour RGB-D images (538 annotated frames, cow-disjoint test set), the network (ConvNeXt-T backbone) reaches \pck{10} 0.905 for landmarks and region IoU 0.70 to 0.91, and runs end to end in 26.4~ms per frame on a T4 GPU, against an 11~fps camera. Removing the region head costs 0.15 \pck{10} (ResNet-18), but a control shows that it mainly adds training signal: wider landmark heatmap targets recover the gain without it. Agreement between the two heads flags unreliable frames without labels. On our own rig at Western Australia's largest dairy farm, the ConvNeXt-T model transfers with ten labelled frames (\pck{10} 0.85), and its regions are carried through depth into a radiometric thermal camera, giving one temperature per region and cow that is stable within a pass.
\end{abstract}

\section{Introduction}
\label{sec:intro}

Precision dairy farming increasingly relies on automated, non-contact measurement of individual cows \cite{halachmi2019}. Many of these measurements concern one part of the body: rump and tail-head temperature around oestrus \cite{perezmarquez2019}, hip width for body weight \cite{song2018}, back posture for lameness \cite{viazzi2014,vanhertem2014,tun2024}, and the rump and tail-head for body condition \cite{spoliansky2016}. Each study defines its region of interest by hand or with a task-specific crop. The region is therefore never measured, reused or validated in its own right, and a new measurement needs a new region-finding method.

A rotary milking parlour is a natural place to measure every cow twice a day: the platform carries each cow past a fixed sensor at a known point in the milking routine. It is also a hard scene for machine vision. The camera sees the cow in the stall together with two neighbours, through rails and cluster hoses, from an oblique view that changes between installations. Colour alone is unreliable on black-and-white coats, and depth alone loses fine landmarks. A front end for such a parlour has to find the right cow, localise anatomy on it and keep up with the platform, on hardware that can sit in the dairy.

Prior work on overhead RGB-D stops short of this. Duan et~al.~\shortcite{duan2025lameness} localise eight dorsal keypoints for lameness features; Ruchay et~al.~\shortcite{ruchay2026bcs} regress a body condition score from the whole frame on the recording our dataset is built on; thermal studies place their regions by hand \cite{perezmarquez2019,salles2016}. None outputs named anatomical regions as masks that a second sensor can reuse.

\begin{figure*}[t]
\centering
\includegraphics[width=\textwidth]{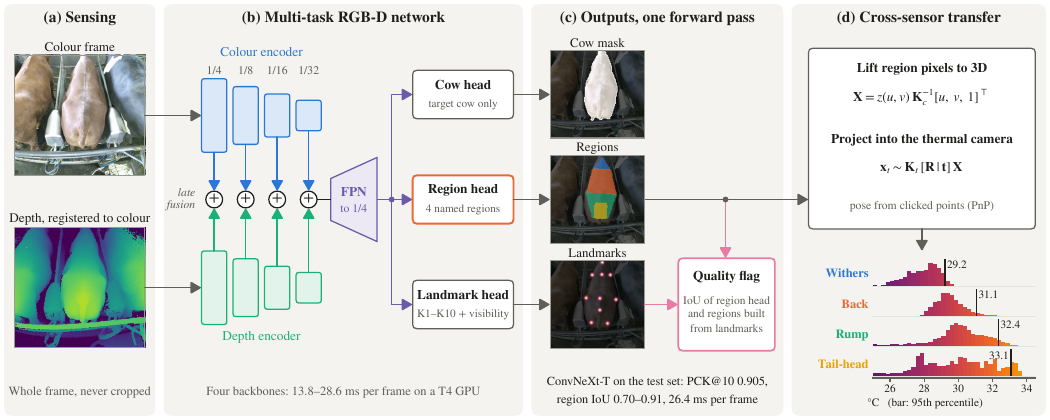}
\caption{The proposed framework. A whole RGB-D frame from the parlour camera passes once through two encoders whose feature pyramids are summed (late fusion) and decoded to stride 4. Three heads give the target-cow mask, ten landmarks with visibility and four named region masks. The agreement between the region head and the regions built from the predicted landmarks serves as a label-free quality flag. The regions are lifted to 3D with the depth behind each pixel and projected into any co-calibrated sensor, here a radiometric thermal camera. (a) to (c) show a test frame from the dataset; (d) shows the temperatures inside each projected region on one farm-rig frame, coloured on the thermal scale, with the 95th percentile that serves as the region's temperature (pixels below 25~$^\circ$C, rails and floor at the region edges, are not drawn).}
\label{fig:framework}
\end{figure*}

We address this gap with one multi-task network, illustrated in \figref{fig:framework}. Its contributions are:
\begin{enumerate}
\item A whole-frame, multi-task RGB-D network (late fusion of two encoders, a shared feature pyramid decoder, and heads for the target-cow mask, ten landmarks with visibility, and four region masks) whose regions come with the depth behind every pixel, so any co-calibrated sensor can receive them.
\item An anatomical region dataset built on the colour and depth images of a public rotary-parlour recording \cite{ruchay2026dataset}: 538 landmark frames, 18 hand-drawn region outlines, 36 within-pass repeat pairs, reviewed target-cow masks and a depth-to-colour calibration.
\item A controlled study on cow-disjoint and session splits, three seeds each: input modality and fusion, head ablations, four backbones including a backbone-specific recipe, and a label-free quality flag from the agreement between the two heads.
\item End-to-end timing on a T4 GPU at batch size one: 13.8 to 28.6~ms per frame, three to six times faster than the camera.
\item An industrial-scale proof of concept on our own rig at Peninsula Downs, Western Australia's largest dairy farm: few-label transfer to a second RGB-D camera and projection of the regions into radiometric thermal images.
\end{enumerate}

\section{Related Work}
\label{sec:related}

\textbf{Dorsal measurement from 3D cameras.} Top-view depth cameras have been used to measure back posture for lameness with accuracy comparable to side-view systems (90\% against 91\%) \cite{viazzi2014}, to classify lameness from consecutive 3D recordings \cite{vanhertem2014} and from backbone height \cite{tun2024}, to predict body weight from hip width (5.2\% error) \cite{song2018}, and to score body condition \cite{spoliansky2016}. These systems derive hand-designed features from a cropped dorsal surface. Spoliansky et~al.~\shortcite{spoliansky2016} report the next cow's head hiding the tail-head as their main failure, a problem that motivates explicit visibility and target-cow isolation in our design.

\textbf{Learned landmarks and scores.} Duan et~al.~\shortcite{duan2025lameness} train a keypoint network on overhead RGB-D images and derive back-curvature features for lameness. Their output is a set of points, not regions, and timing is not reported. Ruchay et~al.~\shortcite{ruchay2026bcs} train a two-stream ResNet-18 on colour and depth from a commercial rotary to regress body condition, reporting 95~ms per frame on their own hardware; their output is one score per frame. Because timings on different hardware are not comparable, we report ours against the camera's frame interval rather than against theirs. Animal pose estimation more broadly follows the heatmap formulation \cite{xiao2018simple,mathis2018deeplabcut}, which we adopt for the landmarks.

\textbf{Regions for thermography.} Infrared studies of dairy cows read temperatures from regions that an operator places on each image: the rump, withers and tail-head warm by 0.30 to 1.20~$^\circ$C in the 48~h before ovulation \cite{perezmarquez2019}, and flank temperature tracks the temperature-humidity index \cite{salles2016}. Placing regions by hand limits such studies to small herds and short trials.

\textbf{RGB-D fusion and multi-task learning.} Fusing depth through a separate encoder improves semantic segmentation over stacking it as an extra input channel \cite{hazirbas2016fusenet}, and jointly learning dense tasks can help each task \cite{kendall2018multitask,prasad2025grasp}. We test both ideas on a small, domain-specific dataset where labels are expensive.

\section{Data and Annotation}
\label{sec:data}

\textbf{Dataset.} We build an anatomical region dataset on the colour and depth images of a public rotary-parlour recording \cite{ruchay2026dataset} (CC BY 4.0). In that recording, a Kinect~v2 about 3~m above a commercial rotary records each cow from behind at an oblique angle as the platform carries it through the field of view. It holds 1,285 passes of 20 frames at about 11~fps from 1,025 cows over four sessions (three afternoons and one morning). Colour frames are $1920\times1080$; depth frames are $512\times424$ in millimetres and are not registered to colour. The recording provides only body condition scores, which we do not use; every landmark, region, mask and calibration in our dataset is our own annotation, described below. We excluded one duplicated pass. Below, the dataset refers to this anatomical region dataset, and the farm rig to our own recordings.

\begin{figure}[t]
\centering
\includegraphics[width=\columnwidth]{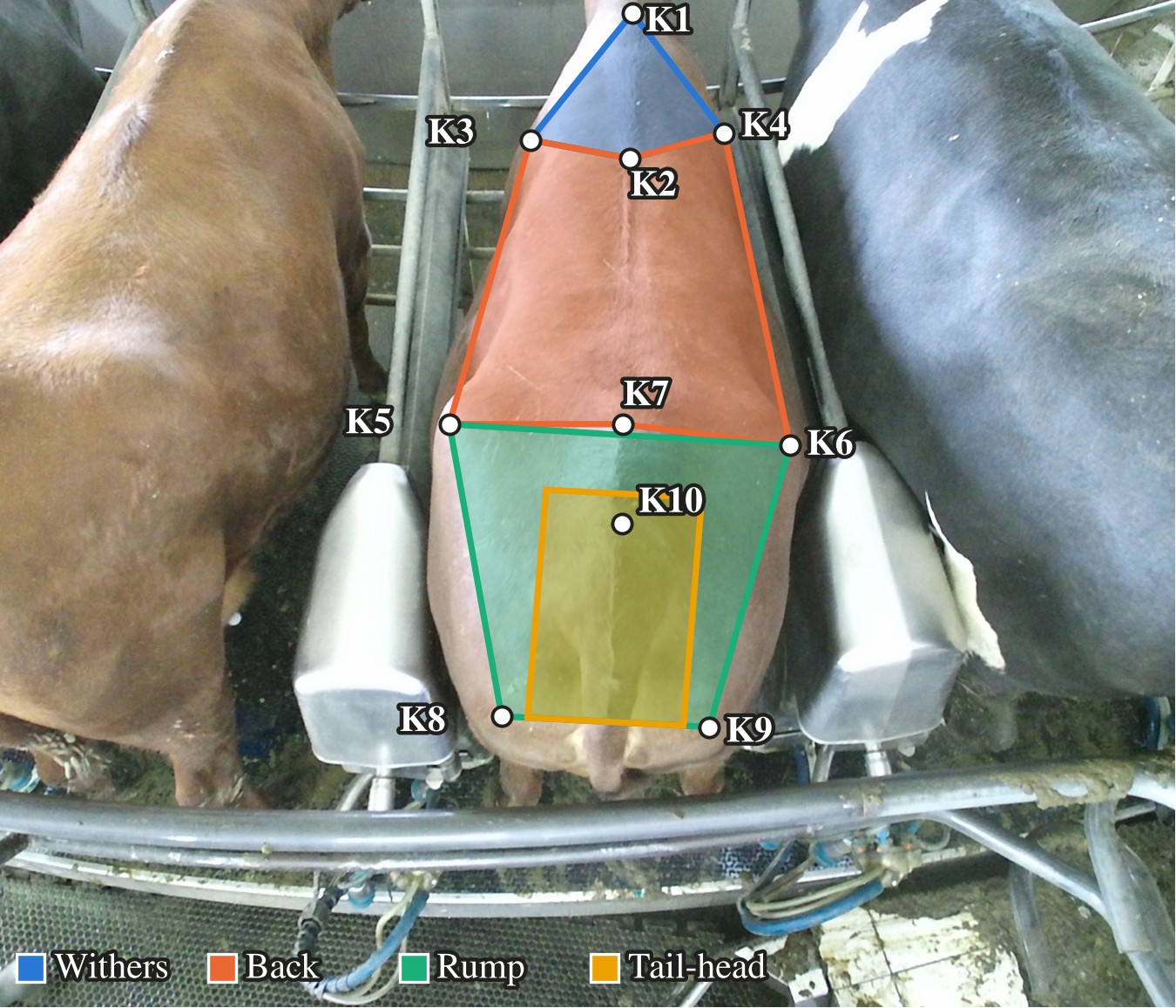}
\caption{Landmarks K1 to K10 and the four rule regions on a test frame: withers (blue), back (orange), rump (green) and tail-head (yellow). The target cow is in the centre stall, between two neighbours.}
\label{fig:scheme}
\end{figure}

\textbf{Landmarks and visibility.} One annotator labelled the middle frame of 538 passes with ten landmarks (\figref{fig:scheme}): the front and rear of the withers (K1, K2), the tops of the shoulder blades (K3, K4), the hooks (K5, K6), the sacrum (K7), the pins (K8, K9) and the tail-head (K10), each with a visibility flag, plus a box around the target cow. Left and right follow the image side. A second frame of 36 passes was labelled again to measure the annotator's repeatability.

\textbf{Rule regions.} The four regions are defined from the landmarks: the withers is the quadrilateral K1, K4, K2, K3; the back is bounded by K3, K2, K4, K6, K7, K5; the rump by the hooks and pins (K5, K6, K9, K8); and the tail-head runs from the pin line to just beyond K10. When K1 is outside the image (4\% to 99\% of frames, depending on the session), it is placed 0.38 hook widths ahead of K2 along the spine. The regions are bounded by bony landmarks rather than the silhouette, so they avoid the steep body edges where thermal readings are least reliable. To check that the rules match anatomy, the annotator drew the four regions freehand on 18 test frames before placing any landmarks on them.

\textbf{Target-cow masks and calibration.} Reference masks of the target cow were produced with SAM~2.1 \cite{ravi2025sam2} prompted by the box and the visible landmarks, and reviewed by eye; none needed editing. Depth was registered to colour with a pose solved by PnP \cite{lepetit2009epnp} from 96 clicked point pairs on 12 frames, which reduced the median reprojection error from 27~px with the nominal Kinect parameters to about 7~px at full resolution.

\textbf{Splits.} Cows are split 70/10/20 into training, validation and test sets, with no cow in two sets, stratified by session and body condition: 406, 58 and 74 frames. A second, session split trains on the afternoons and tests on the morning session (141 frames), whose body condition distribution also differs (31\% of passes at 4.0 or above, against about 1\% in the afternoons).

\textbf{Test frames.} The task defines the target as the cow in the stall nearest the image centre. Test frames whose labelled cow stands in an adjacent stall, more than 300~px (about half a stall) from the image centre, are not scored; they are found from the labels alone. This leaves 69 cow-split test frames (661 labelled landmarks) and 138 session-split test frames.

\section{Method}
\label{sec:method}

\textbf{Input.} The network receives the whole colour frame, scaled by 0.4 to $768\times432$ and padded to $768\times448$, together with depth reprojected into it. The frame is never cropped around the labelled cow: finding the right cow is part of the task.

\textbf{Encoders and fusion.} Colour and depth each pass through their own ImageNet-pretrained encoder \cite{wightman2019timm}; the depth encoder takes one input channel. Their feature maps are summed level by level at strides 4 to 32 (late fusion). The alternative, early fusion, stacks depth as a fourth input channel of one encoder. A feature pyramid decoder \cite{lin2017fpn} with 128 channels merges the levels to stride 4 ($192\times112$ cells).

\textbf{Heads.} Three light convolutional heads read the shared features. The cow head segments the target cow only, so the network must learn to reject the neighbours. The landmark head outputs one heatmap per landmark, trained against a Gaussian of $\sigma=1.5$ cells \cite{xiao2018simple}, and a visibility logit per landmark from globally pooled features; landmarks are decoded by the arg-max with quarter-cell refinement. The region head outputs four masks, trained on the rule regions built from the labelled landmarks. It therefore adds no annotation: it supervises the same landmark labels densely, over every pixel of the dorsal surface, instead of at ten points.

\textbf{Losses.} The total loss is
\begin{equation}
\mathcal{L} = 100\,\mathcal{L}_{\mathrm{hm}} + 0.5\,\mathcal{L}_{\mathrm{vis}} + \mathcal{L}_{\mathrm{reg}} + \mathcal{L}_{\mathrm{cow}} + 0.5\,\mathcal{L}_{\mathrm{cons}},
\end{equation}
where $\mathcal{L}_{\mathrm{hm}}$ is the mean squared heatmap error over labelled landmarks, $\mathcal{L}_{\mathrm{vis}}$ the binary cross-entropy of visibility, and $\mathcal{L}_{\mathrm{reg}}$ and $\mathcal{L}_{\mathrm{cow}}$ the sum of binary cross-entropy and Dice losses for the region and cow masks. The consistency term $\mathcal{L}_{\mathrm{cons}}$ is the Dice loss between the region head and the regions rasterised from the predicted landmarks, applied from epoch 10. The weights were fixed before the experiments and are the same for every run; they were not tuned. Their purpose is to keep the terms on a comparable scale. A heatmap is almost entirely zero, so its mean squared error per cell settles at about $3\times10^{-4}$, only about 15\% below the $3.3\times10^{-4}$ of predicting an all-zero map and more than two orders of magnitude below the mask losses; the factor of 100 brings it to the same order as the cow and region terms (about 0.03 against 0.04 and 0.11 at the end of training for the late-fusion model). Visibility and consistency are auxiliary terms and are halved so that they do not compete with the three main tasks.

\textbf{Label-free quality flag.} At inference, the same comparison gives an agreement score per frame: the mean IoU between the region head's masks and the regions built by rule from the network's own landmarks. The two heads fail in different ways: a landmark error is usually a discrete jump of one heatmap peak (to a rail, the neighbouring cow or the opposite hook), which displaces a whole rule-built region, whereas the region head pools evidence over many pixels and degrades gradually at region boundaries. Because the heads share one encoder, failures with a common cause can still affect both, so low agreement marks a frame whose output should not be trusted, without any label, but high agreement does not guarantee a correct one.

\textbf{Output and projection.} Per frame, the network returns the cow mask, ten landmarks with visibility and four region masks. Because depth is registered to colour, every region pixel has a 3D position in the camera frame. A second camera with known pose and intrinsics receives the regions by projecting those points into its image. We use this in \secref{sec:farm} to read region temperatures from a thermal camera.

\textbf{Training.} All models are trained for 60 epochs with AdamW \cite{loshchilov2019adamw} (learning rate $5\times10^{-4}$, weight decay $10^{-4}$, batch 16), two warm-up epochs then cosine decay, and mixed precision. Augmentation applies horizontal flips with a left-right swap of paired landmarks, scaling by 0.9 to 1.1, rotation by up to $5^\circ$, shifts of up to 30~px and brightness changes. The checkpoint is chosen on validation \pck{10} plus mean region IoU, evaluated every five epochs. ConvNeXt-T \cite{liu2022convnext} was unstable under this shared recipe (three of six runs failed over two learning rates), so it uses its own recipe: encoder learning rate $10^{-4}$ with per-stage decay 0.75, heads at $5\times10^{-4}$, gradient clipping at a norm of 1.0, bf16 precision and ten warm-up epochs.

\section{Experimental Set-up}
\label{sec:setup}

\textbf{Metrics.} Landmark accuracy is the percentage of correct keypoints within 5\% and 10\% of the labelled hook width (\pck{5}, \pck{10}; 10\% is about 37~px at full resolution), and the median error in full-resolution pixels. Regions are scored by IoU against the rule regions built from the labels, and by the share of each predicted region that falls inside the hand-drawn outline. The cow mask is scored by IoU against the reviewed reference mask.

\textbf{Protocol.} Every learned result is the mean and standard deviation over three training seeds, selected on validation and reported on the 69 cow-disjoint test frames (661 labelled landmarks) unless stated. The standard deviation reflects training variation only. To include the variation over test animals, 95\% confidence intervals resample the test cows with replacement (5,000 cluster-bootstrap resamples), recompute each seed's pooled \pck{10} on the resampled frames and average the seeds; a difference between two configurations is resampled on the same cows for both. Input, fusion and head ablations use a ResNet-18 backbone \cite{he2016resnet}; the backbone study compares MobileNetV3-L \cite{howard2019mobilenetv3}, ResNet-18, PVTv2-b1 \cite{wang2022pvtv2} and ConvNeXt-T under the same heads. Within-pass stability is measured without labels on all 20 frames of every pass.

\textbf{Statistics.} Correlations use Spearman's $\rho$, paired comparisons between regions of one cow use the Wilcoxon signed-rank test, and farm results use 4-fold cross-validation by pass.

\textbf{Hardware and timing.} Training uses an NVIDIA A100. Timing uses an NVIDIA T4 at batch size one in FP16, from a frame in host memory to decoded landmarks and masks: upload, GPU pre-processing (depth registration, resizing and normalisation), the network and decoding. Each configuration is warmed up and then timed over at least 500 frames with CUDA synchronisation; we report the median of three fresh sessions.

\section{Results}
\label{sec:results}

\begin{table*}[t]
\centering
\caption{Inputs, fusion and heads (ResNet-18 backbone; mean $\pm$ SD over three seeds; 69 cow-disjoint test frames unless stated). Region IoU is against the rule regions built from the labels; for the landmarks-only model, regions are built by rule from its predicted landmarks.}
\label{tab:ablation}
\small
\setlength{\tabcolsep}{5pt}
\begin{tabular}{lccccccccc}
\toprule
 & & & Median & \multicolumn{5}{c}{IoU} \\
\cmidrule(lr){5-9}
Model & \pck{5} & \pck{10} & error (px) & Withers & Back & Rump & Tail-head & Cow \\
\midrule
RGB only & 0.341 $\pm$ 0.012 & 0.669 $\pm$ 0.013 & 25.7 & 0.65 & 0.85 & 0.87 & 0.72 & 0.94 \\
Depth only & 0.362 $\pm$ 0.003 & 0.701 $\pm$ 0.018 & 24.4 & 0.65 & 0.88 & 0.89 & 0.73 & 0.95 \\
RGB-D, early fusion & 0.338 $\pm$ 0.022 & 0.687 $\pm$ 0.021 & 25.6 & 0.67 & 0.87 & 0.88 & 0.73 & 0.96 \\
\textbf{RGB-D, late fusion (full model)} & \textbf{0.421 $\pm$ 0.014} & \textbf{0.753 $\pm$ 0.007} & \textbf{21.4} & 0.68 & 0.88 & 0.90 & 0.75 & 0.96 \\
\quad without consistency loss & 0.415 $\pm$ 0.015 & 0.745 $\pm$ 0.007 & 21.9 & 0.68 & 0.88 & 0.90 & 0.75 & 0.96 \\
\quad without region head & 0.338 $\pm$ 0.036 & 0.600 $\pm$ 0.026 & 28.3 & 0.41 & 0.72 & 0.76 & 0.55 & 0.96 \\
\quad without landmark head & -- & -- & -- & 0.69 & 0.89 & 0.90 & 0.75 & 0.97 \\
\quad session split (138 frames) & 0.425 $\pm$ 0.017 & 0.752 $\pm$ 0.014 & 22.7 & 0.63 & 0.88 & 0.90 & 0.75 & 0.97 \\
\quad session split, unseen cows only (111) & 0.433 $\pm$ 0.013 & 0.760 $\pm$ 0.017 & 22.5 & 0.63 & 0.88 & 0.90 & 0.75 & 0.97 \\
\bottomrule
\end{tabular}
\end{table*}

\subsection{Accuracy against the human}

The full ResNet-18 model localises the landmarks at \pck{10} 0.753 $\pm$ 0.007 (95\% CI over test cows 0.72 to 0.79; median error 21.4~px at $1920\times1080$), and the best backbone, ConvNeXt-T (\secref{sec:speed}), raises this to 0.905 (0.88 to 0.93). Two kinds of region accuracy are reported. The first is agreement with the rule regions, which are derived from the same landmark labels: back and rump regions reach IoU 0.87 to 0.91 for every RGB-D model with a region head, tail-head 0.73 to 0.77 and withers 0.67 to 0.70 (Tables~\ref{tab:ablation} and~\ref{tab:backbones}). This measures how well the network reproduces the rule, not anatomy.

The second is an anatomical check that does not use the landmarks: freehand outlines of the four regions, drawn before any landmarks were placed. On the 17 outline frames that are scored, a median of 93\% (back) to 99\% (rump and withers) of each rule region lies inside the annotator's freehand outline, and a median of 88\% to 97\% of each region predicted by the ResNet-18 network. They are narrower than the outlines by design, which suits sampling a temperature away from the body edge. The outlines come from the same annotator as the landmarks, so this check is independent of the rule but not of the annotator. Within a pass, the annotator's own repeat labels differ by 2.2\% of hook width for the hook width itself, 4\% to 5\% for the shoulder width and hook-to-pin distance, and 12\% for the distance from the rear of the withers (K2) to the sacrum. K2 and K10 are the landmarks that both the human and the network find hardest.

\begin{figure*}[t]
\centering
\includegraphics[width=\textwidth]{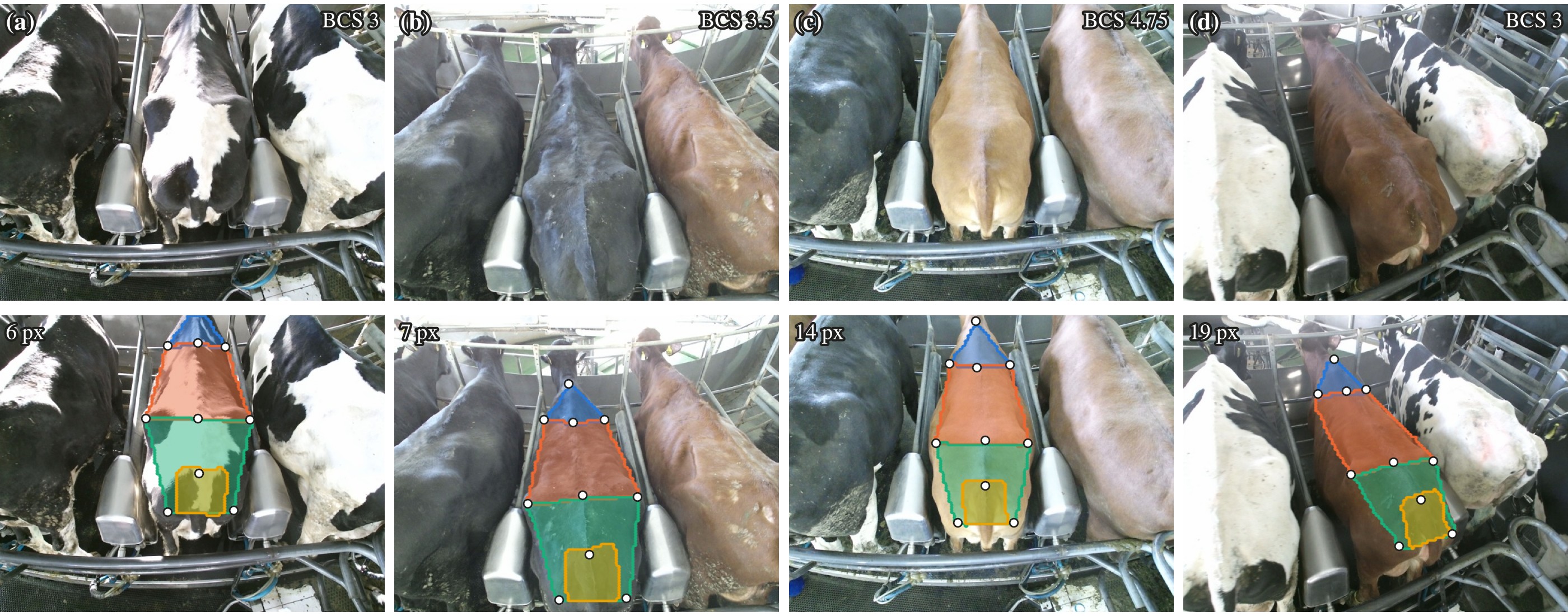}
\caption{Four test cows from the dataset, of different coat colour, size and body condition score (BCS 3 to 4.75), from three sessions (ConvNeXt-T). Top: input colour frames with each cow's BCS. Bottom: predicted regions and landmarks with the frame's median landmark error. (d) is a cow standing at an angle in the stall.}
\label{fig:qual}
\end{figure*}

\figref{fig:qual} shows that the regions follow each cow's own anatomy across coat colours, body sizes and body condition scores. Errors concentrate at the front of the cow and on cows standing at an angle in the stall (\figref{fig:qual}d); the target cow itself is found in every test frame.

\begin{table*}[t]
\centering
\caption{Backbones (RGB-D late fusion, all heads; three seeds; 69 test frames). Times are per frame on a T4 GPU at batch size 1 in FP16, median of three sessions; end to end includes upload, GPU pre-processing and decoding. The camera delivers a frame every 90~ms.}
\label{tab:backbones}
\small
\begin{tabular}{lcccccccc}
\toprule
 & Params & & & Median & IoU & IoU & Network & End to end \\
Backbone & (M) & \pck{5} & \pck{10} & error (px) & back / rump & cow & (ms) & (ms, fps) \\
\midrule
MobileNetV3-L & 6.7 & 0.427 $\pm$ 0.002 & 0.750 $\pm$ 0.007 & 21.5 & 0.89 / 0.90 & 0.97 & 17.9 & 22.4 (45) \\
ResNet-18 & 23.1 & 0.421 $\pm$ 0.014 & 0.753 $\pm$ 0.007 & 21.4 & 0.88 / 0.90 & 0.96 & 10.7 & \textbf{13.8 (73)} \\
PVTv2-b1 & 27.7 & 0.518 $\pm$ 0.004 & 0.843 $\pm$ 0.003 & 17.4 & 0.90 / 0.91 & 0.97 & 25.7 & 28.6 (35) \\
ConvNeXt-T, shared recipe & 56.4 & 0.458 $\pm$ 0.262 & 0.700 $\pm$ 0.330 & 33.1 & 0.88 / 0.90 & 0.97 & 23.5 & 26.4 (38) \\
\textbf{ConvNeXt-T, own recipe} & 56.4 & \textbf{0.645 $\pm$ 0.012} & \textbf{0.905 $\pm$ 0.008} & \textbf{13.3} & \textbf{0.91 / 0.91} & \textbf{0.98} & 23.5 & 26.4 (38) \\
\bottomrule
\end{tabular}
\end{table*}

\subsection{Inputs and fusion}

Depth alone localises the landmarks better than colour alone (\pck{10} 0.701 against 0.669; difference 0.03, 95\% CI 0.00 to 0.06): hooks, pins and the spine are shape, not colour. Feeding both into one encoder gains nothing (0.687), because the colour-pretrained first layer has to absorb a modality it was not trained on. Separate encoders with late fusion gain 0.05 to 0.09 \pck{10} over every single-encoder variant (each 95\% CI excludes zero, \tabref{tab:diff}) and reduce the seed-to-seed spread (\tabref{tab:ablation}). Region IoU is nearly the same for all RGB-D variants, so the gain from late fusion lies in the landmarks.

\subsection{Multi-task design and quality flag}

\begin{table}[t]
\caption{Paired differences in \pck{10} on the 69 test frames, with 95\% confidence intervals from a bootstrap over test cows (mean of three seeds per configuration; ResNet-18 unless stated). Point estimates come from re-scoring the saved models and agree with Tables~\ref{tab:ablation} and~\ref{tab:backbones} within 0.005.}
\label{tab:diff}
\small
\setlength{\tabcolsep}{4pt}
\begin{tabular}{lcc}
\toprule
Comparison & $\Delta$\pck{10} & 95\% CI \\
\midrule
Late vs early fusion & 0.068 & [0.040, 0.095] \\
Late fusion vs depth only & 0.054 & [0.031, 0.078] \\
Late fusion vs colour only & 0.086 & [0.061, 0.110] \\
Depth only vs colour only & 0.032 & [0.003, 0.062] \\
With vs without region head & 0.155 & [0.124, 0.185] \\
No region head: loss $\times$10 vs $\times$1 & 0.049 & [0.027, 0.071] \\
No region head: $\sigma$ 3 vs 1.5 & 0.219 & [0.189, 0.248] \\
No region head, $\sigma$ 3, vs full model & 0.065 & [0.044, 0.085] \\
With vs without consistency loss & 0.013 & [0.001, 0.024] \\
ConvNeXt-T vs ResNet-18 & 0.150 & [0.118, 0.183] \\
ConvNeXt-T vs PVTv2-b1 & 0.062 & [0.037, 0.087] \\
\bottomrule
\end{tabular}
\end{table}

Removing the region head lowers landmark \pck{10} from 0.753 to 0.600 (difference 0.155, 95\% CI 0.12 to 0.19) and increases the median error from 21.4 to 28.3~px. The region targets are built from the same landmark labels, so the gain cannot come from new information; it comes from how the information is delivered. A heatmap is almost entirely zero, and its loss stays close to that of predicting nothing at all (\secref{sec:method}), whereas the region loss gives every pixel of the dorsal surface a target. Two controls without a region head separate signal from shape (\tabref{tab:diff}). A ten times stronger heatmap loss gains only 0.05 (\pck{10} 0.650 $\pm$ 0.022), but heatmap targets twice as wide ($\sigma=3$ cells) reach 0.819 $\pm$ 0.019 (median error 18.5~px), above the full model by 0.065 (95\% CI 0.04 to 0.09). The region head therefore helps the landmarks mainly by adding dense training signal, not anatomical information that the landmarks lack: on 406 training frames a sharp heatmap target is too sparse to learn from, and either the regions or a wider target supplies the missing signal. What the region head adds beyond this is the regions themselves, segmented directly (back and rump IoU 0.88 and 0.90, against 0.86 and 0.87 when built by rule from the $\sigma=3$ landmarks), and the second opinion behind the quality flag. Whether a wider target also improves the full model was not tested. The reverse does not hold: a regions-only model segments the regions as well as the full model, so the landmark head is not needed for the regions. Removing the consistency loss lowers \pck{10} only slightly (0.753 against 0.745; paired difference 0.01, 95\% CI 0.00 to 0.02): the shared multi-task design, not the explicit consistency term, provides most of the benefit.

\begin{figure}[t]
\centering
\includegraphics[width=\columnwidth]{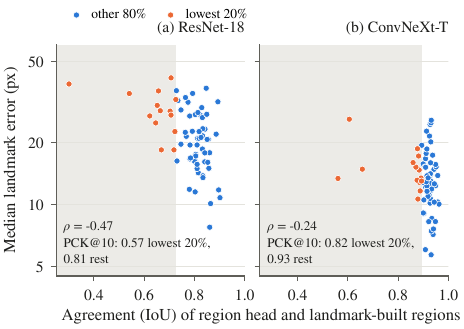}
\caption{Label-free quality flag on the 69 test frames (seed 0). Frames whose two heads disagree are the frames with large landmark errors; the shaded band is the lowest 20\% of agreement. The relation is clear for ResNet-18 and weaker for ConvNeXt-T, whose errors are smaller.}
\label{fig:flag}
\end{figure}

The agreement between the two heads is nevertheless useful at inference (\figref{fig:flag}). With ResNet-18 it correlates with the median landmark error of the frame (Spearman $\rho=-0.47$ for the seed shown, $-0.28$ to $-0.47$ over three seeds), and the 20\% of frames with the lowest agreement have a mean \pck{10} of 0.59 against 0.80 for the rest (means over seeds). The flag weakens as the model improves, because its errors become smaller and rarer: with the final ConvNeXt-T model, $\rho$ is $-0.08$ to $-0.24$ and the lowest 20\% reach \pck{10} 0.84 against 0.92. On the 52 farm frames (\secref{sec:farm}) the flag holds for the transferred models: $\rho=-0.46$ ($p<0.001$) for the unchanged dataset model and $-0.35$ ($p=0.01$) after fine-tuning on 39 frames, where the lowest 20\% reach \pck{10} 0.43 and 0.83 against 0.58 and 0.94 for the rest. The flag is therefore most useful where the model is weakest, such as a new installation before fine-tuning.

\subsection{The right cow}

Against the reviewed reference masks, a depth-only cut of the stall cow reaches a mean IoU of 0.74, and SAM~2 prompted from depth 0.77: both merge touching neighbours. The network's cow head reaches 0.96 with ResNet-18 and 0.98 with ConvNeXt-T. Trained on whole frames, it learns which cow is the target from the stall layout rather than from a crop.

\subsection{Sessions and stability}

Trained on the three afternoon sessions and tested on the morning session, the full model reaches \pck{10} 0.752 $\pm$ 0.014, the same as on the cow split, despite the different body condition mix (\tabref{tab:ablation}). This split is not cow-disjoint: 27 of the 138 morning test frames show cows that also appear in afternoon training passes. On the 111 frames of cows never seen in training, \pck{10} is 0.760 $\pm$ 0.017 (95\% CI 0.73 to 0.79), so the result does not rest on cows the model has seen. Without labels, we compare the ResNet-18 model's shape descriptors (hook width, and ratios to it) from two frames about one second apart in 199 passes. They change by a median of 1.4\% (hook width) to 8.0\% (pin width). For four of the six descriptors the annotator also measured, the model is steadier than the annotator's repeat labels (for example, withers to sacrum 4.6\% against 12.1\%); shoulder and pin widths are less stable than the human. Absolute widths converted to millimetres through the Kinect depth do not repeat for the same cow across dates (ICC 0.02 to 0.12 over 215 to 218 cows), so we use relative shape only.

\subsection{Backbones and speed}
\label{sec:speed}

\tabref{tab:backbones} compares four backbones. With its own recipe, ConvNeXt-T is the most accurate (\pck{10} 0.905 $\pm$ 0.008, median error 13.3~px) and converges on all three seeds; under the shared recipe it failed in three of six runs over two learning rates, collapsing or never converging. The gain therefore depends on optimisation as much as on capacity: on 406 training frames, a large pretrained encoder needs a lower, layer-wise learning rate and gradient clipping. PVTv2-b1 is the most accurate backbone under the shared recipe (0.843). MobileNetV3-L matches ResNet-18 (0.750 against 0.753) with 3.5 times fewer parameters but is not faster on a T4, whose throughput favours the dense convolutions of ResNet-18.

Moving depth registration from the CPU (62~ms per frame) to the GPU (2.5 to 2.9~ms) makes the network the main cost. End to end, a frame takes 13.8~ms with ResNet-18 (73~fps) and 26.4~ms with ConvNeXt-T (38~fps; 95th percentile 28.1~ms), against a camera frame interval of about 90~ms. GPU and CPU pre-processing give the same landmarks (median difference 0~px on ten test frames). Every backbone thus processes every frame with a margin of three to six times, which leaves room for tracking, a second camera or a smaller embedded GPU.

\begin{figure}[t]
\centering
\includegraphics[width=\columnwidth]{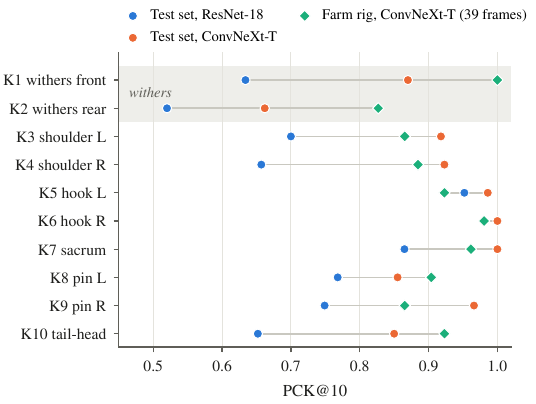}
\caption{Per-landmark \pck{10} (test set: mean of three seeds; farm rig: 4-fold cross-validation). On the test set the rear of the withers (K2) is the hardest landmark for both backbones; on the farm rig's steeper view the same ConvNeXt-T backbone finds both withers landmarks more reliably.}
\label{fig:perkp}
\end{figure}

\figref{fig:perkp} breaks accuracy down by landmark. Hooks and sacrum are the easiest. The front of the cow is the hardest on this dataset: the rear of the withers (K2) reaches \pck{10} 0.66 even with ConvNeXt-T, and the withers region has the lowest IoU. We return to this in \secref{sec:farm}.

\begin{figure*}[t]
\centering
\includegraphics[width=\textwidth]{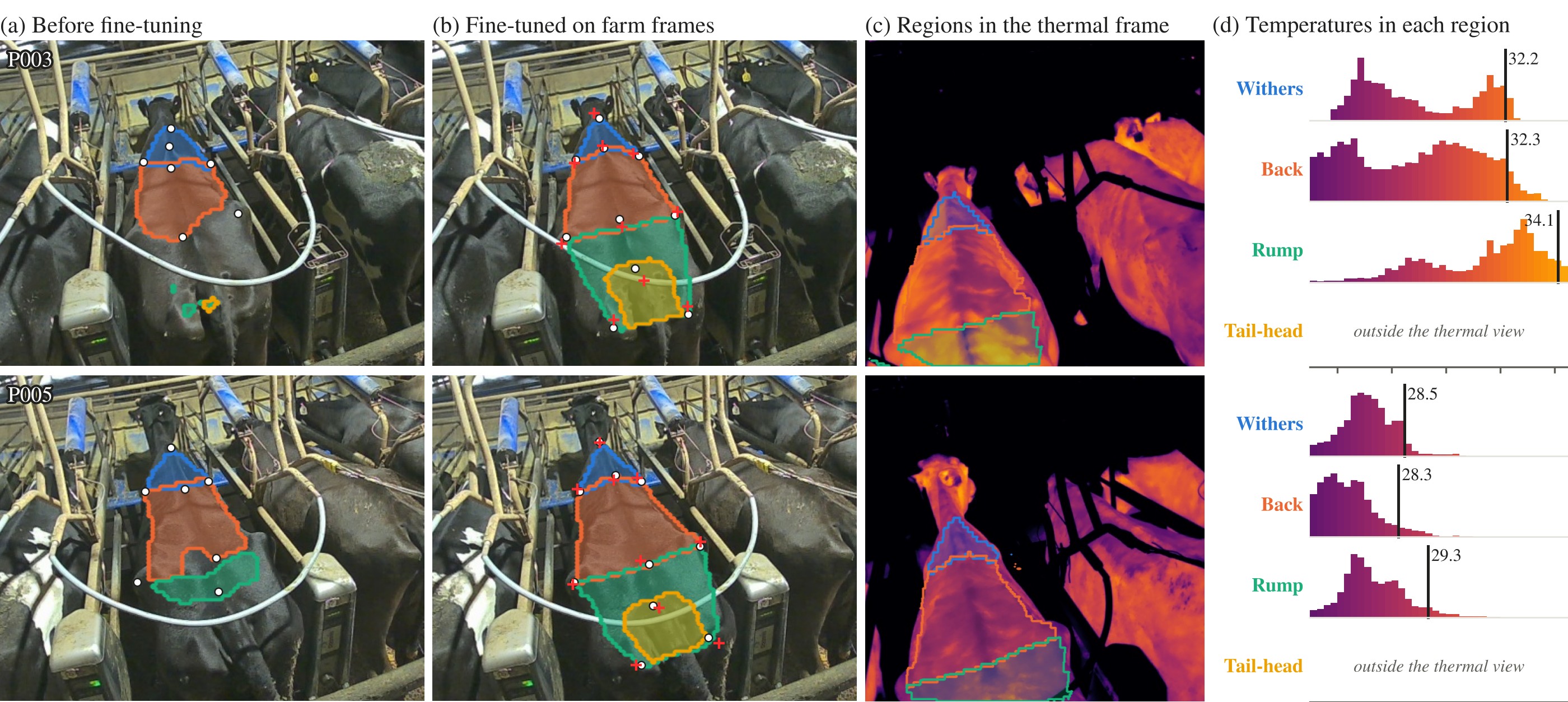}
\caption{Two farm passes held out from the model shown (ConvNeXt-T). (a) The model trained on the dataset, unchanged. (b) After fine-tuning on 39 farm frames; red crosses are the labels. (c) The fine-tuned regions lifted to 3D with the RGB-D depth and projected into the thermal frame of the same moment. (d) Temperatures inside each projected region, coloured on the thermal scale, with the 95th percentile marked. In these passes the tail-head lies outside the thermal camera's narrower field of view; a strictly overhead mount with matched fields of view will close this gap.}
\label{fig:farm}
\end{figure*}

\subsection{Farm proof of concept at Peninsula Downs}
\label{sec:farm}

\textbf{Rig and data.} We recorded a 9-minute milking session on the commercial rotary parlour at Peninsula Downs, Western Australia's largest dairy farm, with a Basler Stereo mini RGB-D camera (colour and pixel-aligned depth, $1280\times800$) and a FLIR A70 radiometric thermal camera ($640\times480$, emissivity 0.95), both mounted above and behind the stalls in a view that is steeper and wider than the Kinect's. There was no calibration target and no shared clock: the clocks were aligned on a bracket knock seen by both cameras, and the thermal pose was solved from 101 clicked correspondences on ten frame pairs by PnP with a one-dimensional search over the focal length. The solution (focal length 681~px, a 50$^\circ$ horizontal field of view against the lens's nominal 51$^\circ$) reprojects with a median error of 3.5~px when each frame pair is held out, under 1~cm at the cow. We extracted 52 passes of 20 frames and labelled the frame with the rump centred in each pass. Farm frames are scaled using depth only, so that a cow's back appears the same size as in training (scale 0.637).

\begin{table}[t]
\centering
\caption{Transfer to the farm rig (ConvNeXt-T; 52 labelled frames, 4-fold cross-validation by pass; errors in farm pixels, pooled over the 52 held-out frames). $\pm$ is the SD over the four folds, and for 10 and 20 frames also over three random draws of the training frames. Every row except the last starts from the model trained on the dataset.}
\label{tab:farm}
\small
\setlength{\tabcolsep}{2pt}
\begin{tabular}{lcccc}
\toprule
 & & Median & \multicolumn{2}{c}{IoU} \\
\cmidrule(lr){4-5}
Setting & \pck{10} & error (px) & Back & Rump \\
\midrule
Zero-shot (no farm labels) & 0.545 $\pm$ 0.045 & 16.7 & 0.55 & 0.41 \\
Fine-tuned, 10 frames & 0.846 $\pm$ 0.028 & 9.1 & 0.86 & 0.86 \\
Fine-tuned, 20 frames & 0.890 $\pm$ 0.027 & 8.2 & 0.87 & 0.88 \\
Fine-tuned, 39 frames & \textbf{0.913 $\pm$ 0.040} & \textbf{7.6} & \textbf{0.88} & \textbf{0.89} \\
ImageNet start, 39 frames & 0.533 $\pm$ 0.026 & 17.5 & 0.77 & 0.82 \\
\bottomrule
\end{tabular}
\end{table}

\begin{figure}[t]
\centering
\includegraphics[width=\columnwidth]{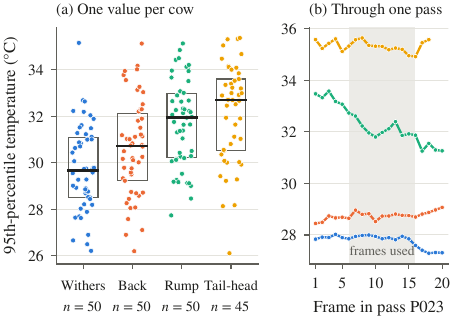}
\caption{Region temperatures on the farm rig. (a) One 95th-percentile value per cow and region (box: quartiles; line: median). (b) The same value frame by frame through one pass; shaded frames are used.}
\label{fig:p95}
\end{figure}

\textbf{Fine-tuning protocol.} The 52 labelled passes are split by pass into four folds of 13. Each fold is tested with a model fine-tuned on the passes of the other three folds (39 frames), or on 10 or 20 of them drawn at random, three times per fold. Fine-tuning starts from the model trained on the dataset (or from ImageNet for the last row of \tabref{tab:farm}) and runs for 40 epochs with batch 8, the same augmentation and the ConvNeXt-T recipe: heads at $10^{-4}$, encoders at $2\times10^{-5}$ with per-stage decay 0.75, gradient clipping and five warm-up epochs, without the consistency term. There is no farm validation set, so the last epoch is used. Errors are pooled over the 52 held-out frames.

\textbf{Transfer.} Unchanged, the model trained on the dataset finds the stall cow on the new rig and places its landmarks at \pck{10} 0.545, but its rump and tail-head regions are incomplete (\tabref{tab:farm}, \figref{fig:farm}). Fine-tuning on ten labelled farm frames raises \pck{10} to 0.846 and the back and rump IoU to 0.86; with 39 frames it reaches 0.913 (median error 7.6~px). Across folds, the fine-tuned results vary by 0.03 to 0.04 (SD), and the choice of the 10 or 20 training frames by about 0.01. Trained from ImageNet on the same 39 frames, the network reaches only 0.533, about the level that the dataset-trained model achieves with no farm labels at all: training on the dataset is worth about 40 labelled farm frames here. On this steeper view the withers is no longer the weak point: the front and rear of the withers reach \pck{10} 1.00 and 0.83 (\figref{fig:perkp}), which supports mounting the camera more directly overhead.

\textbf{One temperature per region and cow.} For each frame we take the 95th percentile of the thermal pixels inside each projected region, using the fold model that did not see that pass, and then the median over frames 6 to 16 of the pass, keeping regions with at least 50 thermal pixels. Over 50 passes, this value changes by only 0.11 to 0.29~$^\circ$C (SD) within a pass while the cow crosses the field of view, against 1.85 to 2.16~$^\circ$C between cows (\figref{fig:p95}). This is stability, not accuracy: the recording had no radiometric reference in the scene, so whether the values are correct in absolute terms is not tested here. The withers and back lie fully inside the narrower thermal view, but on average only 60\% of the rump and 29\% of the tail-head do, and the tail-head has fewer than 50 thermal pixels in 18\% of frames, so 45 of the 50 passes give a tail-head value. For these two regions the value describes the visible part only. Within a pass it rises slightly as more of the region comes into view (Spearman $\rho$ 0.25 for the rump and 0.36 for the tail-head, about 0.15 and 0.24~$^\circ$C per additional tenth of the region), whereas across cows it does not depend on how much is visible ($\rho$ 0.02 and 0.20, $p>0.1$). The regions differ systematically within the same cow: the rump is 1.9~$^\circ$C warmer than the withers and 1.1~$^\circ$C warmer than the back (Wilcoxon $p<10^{-9}$), consistent with the regional differences that hand-placed studies report \cite{perezmarquez2019}.

\FloatBarrier
\section{Discussion and Limitations}
\label{sec:discussion}

\textbf{Fusion and supervision matter more than modality.} Depth alone already beats colour, but the decisive choices concern fusion and supervision: separate encoders, and enough training signal for the landmarks, whether it comes from a dense region task or from a wider heatmap target. The extra signal is cheap at inference: a wider target costs nothing, and the region head adds one convolutional head to a shared decoder.

\textbf{Real-time margin.} The full pipeline runs at 35 to 73~fps on a 70~W inference GPU. The margin over an 11~fps camera matters on a rotary, where every cow passes once per milking and a missed frame cannot be recovered; it also leaves room for an embedded GPU in the parlour, which we have not yet measured.

\textbf{Limitation: camera placement.} The main limitation is the viewing geometry, not the network. On the dataset's oblique Kinect view, the withers is the weak point: K2 is the least repeatable human landmark, K1 is often outside the frame, and the low angle foreshortens the front of the cow; on our steeper farm view, the same landmarks are among the easiest. On the farm rig, the thermal camera is narrower than the colour camera, so only about 60\% of rump pixels and 29\% of tail-head pixels fall inside the thermal frame, and the tail-head is often missing (\figref{fig:farm}). A strictly overhead mount with matched colour, depth and thermal fields of view removes both problems: it views the whole back without foreshortening and keeps every region inside every camera.

\section{Conclusion}
\label{sec:conclusion}

A single RGB-D network can find the target cow in a rotary parlour, localise ten dorsal landmarks and segment four named regions from the whole frame, faster than the camera delivers frames. Late fusion and dense supervision of the landmarks are the decisive design choices, and agreement between the heads flags unreliable frames without labels. Because each region carries its depth, a second sensor can receive it: on our own rig at a commercial dairy, ten labelled frames adapt the model, and the regions yield one temperature per region and cow that is stable within a pass. The next step is a strictly overhead rig with matched colour, depth and thermal fields of view, followed by multi-session farm recordings to test whether region temperatures carry physiological signal.

\section*{Acknowledgements}
We thank Peninsula Downs and its staff for access to the parlour. This work was supported by the Australian Government's Australia's Economic Accelerator (AEA) Ignite Program under Grant No.~IG250201257.

\bibliographystyle{named}
\bibliography{refs}

\end{document}